\documentclass{ifacconf}

\usepackage[utf8]{inputenc}
\usepackage{natbib}
\usepackage{amsmath}
\usepackage{amssymb}
\usepackage{algorithm}
\let\classAND\AND
\let\AND\relax
\usepackage{algorithmic}

\let\AND\classAND
\usepackage{graphicx}
\usepackage{booktabs}
\usepackage{multirow}
\usepackage[flushleft]{threeparttable}
\usepackage{mathtools}

\usepackage{url}
\usepackage{bm}
\usepackage{proof}
\newtheorem{theorem}{Theorem}
\newtheorem{lemma}[theorem]{Lemma}

\makeatletter
\newenvironment{breakablealgorithm}
  {% begin code
   \begin{center}
     \refstepcounter{algorithm}% New algorithm
     \hrule height.8pt depth0pt \kern2pt% \@fs@pre for \@fs@ruled
     \renewcommand{\caption}[2][\relax]{% Make a new caption command
       {\raggedright\textbf{Algorithm~\thealgorithm} ##2\par}%
       \ifx\relax##1\relax % if no label given
         \addcontentsline{loa}{algorithm}{\protect\numberline{\thealgorithm}##2}%
       \else % if label given
         \addcontentsline{loa}{algorithm}{\protect\numberline{\thealgorithm}##2}%
         \protected@write\@auxout{}{\string\newlabel{##1}{{\thealgorithm}{\thepage}}}%
       \fi
       \kern2pt\hrule\kern2pt
     }
  }{% end code
   \kern2pt\hrule\relax% \@fs@post for \@fs@ruled
   \end{center}
  }
\AtBeginEnvironment{algorithmic}{\let\AND\algoAND}
\floatname{algorithm}{Algorithm}

\makeatother

\begin{document}
\begin{frontmatter}

\title{Efficient Verification of Neural Control Barrier Functions with Smooth Nonlinear Activations} 

\thanks[footnoteinfo]{The work is supported by the National Natural Science Foundation of China under Grant 62373239, 62333011, 62461160313, and the State Key Laboratory of Space Intelligent Control under grant HTKJ2025KL502025. (Corresponding author: Liang Xu.)}

\author[First]{Jun Zhang} 
\author[Second]{Haibo Zhang} 
\author[Third]{Chun Liu}
\author[Fourth]{Xiaofan Wang}
\author[Fifth]{Liang Xu}

\address[First]{School of Mechatronic Engineering and Automation, Shanghai University, Shanghai, China (e-mail: jun\_zhang@shu.edu.cn).}
\address[Second]{Beijing Institute of Control Engineering, Beijing, China (e-mail: zhanghb502@163.com).}
\address[Third]{School of Mechatronic Engineering and Automation, Shanghai University, Shanghai, China (e-mail: chun\_liu@shu.edu.cn).}%PIN 156764
\address[Fourth]{School of Mechatronic Engineering and Automation, Shanghai University, Shanghai, China (e-mail: xfwang@shu.edu.cn).}
\address[Fifth]{School of Future Technology, Shanghai University, Shanghai, China (e-mail: liang-xu@shu.edu.cn).}

\begin{abstract}
Formal verification of neural control barrier functions (NCBFs) remains challenging, especially for neural networks with nonlinear activations like \(\tanh\). Existing CROWN-based methods rely on conservative linear relaxations for Jacobian bounds, limiting scalability. We propose LightCROWN, which computes tighter Jacobian bounds by exploiting the analytical properties of activation functions. Experiments on nonlinear control systems including the inverted pendulum, Dubins car, and planar quadrotor demonstrate that LightCROWN improves verification success rates up to 100\%, while enhancing speed and scalability. Our approach provides a generalizable improvement for CROWN-based frameworks, enabling more efficient verification of complex NCBFs. The code can be found at \url{github.com/Autonomous-Systems-and-Control-Lab/verify-neural-CBF}.
\end{abstract}


\end{frontmatter}
%===============================================================================

\section{Introduction}
Safety is a fundamental property of control systems, ensuring that a system's state remains within a prescribed safe set. In safety-critical applications—such as autonomous driving, unmanned aerial vehicles, and robotic manipulation—violating safety constraints can lead to irreversible consequences, including loss of human life or the destruction of critical infrastructure~\cite{ames2019control,JQRR202303009,JQRR202501009}. Consequently, ensuring safety is a non-negotiable requirement for the design and deployment of autonomous systems.

Control Barrier Functions (CBFs)~\cite{ames2019control} provide a formal mechanism to certify safety. Given a safe set $\mathcal{C} \subset \mathbb{R}^n$ defined by a function $h(x) \le 0$, a valid CBF $h$ ensures the \emph{forward invariance} of $\mathcal{C}$ via Nagumo's theorem~\cite{nagumo1942lage}. This property allows for the synthesis of controllers that provably keep system trajectories inside $\mathcal{C}$ for all future time. When $h(x)$ is given analytically, verifying that it satisfies the required CBF conditions is often straightforward.

To handle the complex and irregular safe sets found in realistic systems, CBFs have been extended to Neural Control Barrier Functions (NCBFs), which represent $h(x)$ with a neural network~\cite{Dawson2023safe}. While NCBFs offer significantly enhanced expressive power, this flexibility introduces a formidable challenge: verification~\cite{clark2024semi}. For an NCBFs to be trustworthy, it must provably satisfy the forward invariance condition across the state space. Without a verification guarantee, deploying an NCBF-based controller is unacceptably risky. This paper addresses the fundamental problem of how to verify NCBFs with high efficiency and accuracy.

Recent efforts to directly verify the forward invariance of learned NCBFs have explored several directions, including counterexample-guided falsification using SMT solvers~\cite{gao2013dreal,peruffo2021automated, abate2020formal}, mixed-integer programming(MIP)~\cite{zhao2022verifying,huang2025dynamic}, Lipschitz neural networks~\cite{tayal2024learning}, and linear bound propagation based on the CROWN framework~\cite{mathiesen2022safety,wang2023simultaneous, chen2024verification}. SMT and MIP based methods provide exact guarantees but suffer from limited scalability, making them impractical for larger neural networks. Approaches relying on Lipschitz networks improve tractability but often impose restrictive architectural constraints that limit expressiveness. In contrast, CROWN-based linear bound propagation achieves a favorable balance, offering scalability to large neural networks while retaining flexibility in network design. However, although these linear bound methods scale well~\cite{zhang2018efficient,xu2020automatic,yang2024lyapunov}, their effectiveness diminishes when computing the inner product between the Jacobian of the neural CBF and nonlinear dynamics~\cite{wang2023simultaneous}, since the resulting bounds lose linear structure and interval arithmetic leads to overly conservative estimates. Recent work using McCormick-style affine relaxations provides an alternative route to reduce conservatism~\cite{vertovec2025scalable}, but such methods introduce additional propagation parameters and engineering choices. LightCROWN aims to be a simpler, parameter-free enhancement that directly exploits analytic derivative properties where available.

Therefore, while recent work by Hu et al.~\cite{hu2024verification} introduced an efficient verification framework for ReLU-based NCBFs, its applicability is limited to non-smooth activations and cannot be extended to continuous-time systems that require smooth such as tanh or sigmoid. Existing methods that support these smooth activations, such as Branch-and-Bound Verification scheme (BBV)~\cite{wang2024simultaneous}, rely on generic and conservative linear approximations. This approach fails to leverage the specific analytical properties of these functions, leading to looser Jacobian bounds and significantly degrading verification efficiency and precision. For example, the auto\_LiRPA implementation (see \cite{xu2020automatic}) includes specialized handlers such as \texttt{BoundTanhGrad} that exploit properties of the tanh derivative. However, auto\_LiRPA and similar CROWN implementations perform layer-wise linear relaxations to obtain activation and derivative bounds across all layers~\cite{xu2020automatic}. In contrast, LightCROWN avoids a final layerwise linear relaxation: for activations like $\tanh$ we compute the derivative extremum on the pre-activation interval analytically and use these exact derivative bounds directly in the backward Jacobian propagation (see Eq.~\eqref{eq:jacobian_bounds_lightcrown}), yielding tighter Jacobian intervals in practice.

To bridge this critical gap, we propose LightCROWN. Our method draws inspiration from the efficiency of framework by~\cite{hu2024verification} but extends it to smooth activations. The core innovation is to replace the generic linear approximations with tighter, property-driven bounds that leverage the specific analytical properties of these functions (e.g., the monotonicity of their derivatives). This approach enables us to achieve significantly tighter Jacobian bounds than generic methods like BBV~\cite{wang2024simultaneous}, resulting in a more efficient and accurate verification process for a broader range of NCBFs.

%Our Contributions
Our contributions are as follows:
\begin{itemize}
    \item We propose LightCROWN, a novel method for tightening verification bounds within the CROWN framework. Our key technical contribution is the replacement of generic linear relaxations of activation function derivatives with tighter bounds derived directly from their analytical properties (e.g., derivative monotonicity).
    \item We apply LightCROWN to NCBFs verification and demonstrate better performance. The tighter bounds lead to significant improvements in both verification success rate (by up to 100\%) and scalability, enabling the certification of more complex and expressive NCBFs.
    \item Beyond NCBFs verification, LightCROWN provides a general mechanism for precise Jacobian estimation. These estimates directly apply to sensitivity analysis, robustness verification, and Lipschitz constant estimation, highlighting its broader utility in learning-enabled control and safety-critical systems.
\end{itemize}

%paper organization
The remainder of this paper is organized as follows. 
Section~\ref{sec:preliminaries_problem} provides background on control barrier functions and formal verification preliminaries. 
Section~\ref{sec:verification_framework} introduces the proposed \textsc{LightCROWN} framework in detail. 
Section~\ref{sec:experiments} reports experimental on multiple nonlinear control benchmarks. 
Finally, Section~\ref{sec:conclusion} summarizes the contributions and discusses potential directions for future work.

\section{Preliminaries and Problem Formulation}
\label{sec:preliminaries_problem}

In this section, we first present the Preliminaries, which introduce the notion of system dynamics, the formal definition of safety, the construction of a safety set via a neural control barrier function, and the role of the Lie derivative in ensuring forward invariance. Once these foundations are laid, we turn to the Problem Formulation.

% \subsection{Preliminaries}
% \textbf{System Dynamics.}
\subsection{Preliminaries}
Consider a nonlinear control system
\begin{equation}
  \dot{x} = f(x,u), \quad x \in D \subset \mathbb{R}^n,\; u \in U \subset \mathbb{R}^k,
  \label{eq:dynamics}
\end{equation}
where \(D\) is the state space and \(U\) the set of admissible inputs. 

% \textbf{Safety Definition.}
Formally, the system is said to be safe with respect to a set \(C \subseteq D\) if the following holds:
\begin{equation}
  x(0) \in C \quad \Rightarrow \quad x(t) \in C,\quad \forall t \ge 0.
\end{equation}
That is, once the system state starts in the safe set \(C\), it must remain in \(C\) for all future time. This property is referred to as forward invariance of the set \(C\).

% \textbf{Safety Set via NCBF}
We encode the safe set \(C\) via a scalar-valued function \(h(x):\mathbb{R}^n \to \mathbb{R}\), known as a NCBFs. Specifically, the safe set is defined as:
\begin{align}
  C &= \left\{ x \in D \,\middle|\, h(x) \leq 0 \right\}, \\
  \partial C &= \left\{ x \in D \,\middle|\, h(x) = 0 \right\}, \\
  \operatorname{Int}(C) &= \left\{ x \in D \,\middle|\, h(x) < 0 \right\}.
\end{align}
States within \( \operatorname{Int}(C) \) are considered strictly safe, and the boundary \( \partial C \) marks the edge of the safety region.

% \textbf{Forward Invariance Condition}
To ensure that trajectories starting in \(C\) remain in \(C\), we require the set \(C\) to be forward invariant under the system dynamics \eqref{eq:dynamics}. This is guaranteed if there exists a control input \(u \in U\) such that the following condition holds for all \(x \in C\):
\begin{equation}
  \mathcal{L}_f h(x) + \alpha(h(x)) \leq 0,
  \label{eq:cbf_condition}
\end{equation}
where \(\alpha(\cdot)\) is an extended class-\(\mathcal{K}\) function and \(\mathcal{L}_f h(x)\) denotes the Lie derivative of \(h\) along the vector field \(f\). An extended class-\(\mathcal{K}\) function is a continuous, strictly increasing function \(\alpha:\mathbb{R}\to\mathbb{R}\) with \(\alpha(0)=0\). This condition implies that when the system state approaches the boundary of the safe set, appropriate control inputs can be chosen to remain in \(C\).

% \textbf{Lie Derivative and Gradient Dependency}
The Lie derivative appearing in \eqref{eq:cbf_condition} is defined as:
\begin{equation}
  \mathcal{L}_f h(x)
  = \nabla h(x)^\top f(x,u)
  = \frac{\partial h}{\partial x} f(x,u).
  \label{eq:lie}
\end{equation}
This quantity captures the instantaneous rate of change of the barrier function along the system's trajectory. Crucially, the computation of \(\mathcal{L}_f h(x)\) depends on the gradient of the NCBFs, i.e., the Jacobian \(\nabla h(x)\). As such, accurately estimating \(\nabla h(x)\) is essential for verifying whether the CBF condition is satisfied.

% \textbf{Gradient of the Neural Barrier Function}
In general, NCBFs \(h(x)\) are implemented as feedforward neural networks (FCNNs) like~\cite{hu2024verification,wang2024simultaneous}. Unlike prior work such as ~\cite{hu2024verification}, which primarily employs piecewise-linear activations, we adopt continuously differentiable nonlinear activations (e.g., $\tanh$). Suppose \(h(x)\) is implemented as an \(L\)-layer feedforward neural network:
\begin{equation}
\begin{aligned}
  h(x) &= M_\theta(x) \\
       &= W_L\,\sigma_{L-1}\bigl(\cdots\sigma_1(W_1 x + b_1)\cdots\bigr) + b_L,
  \label{eq:ncbf}
\end{aligned}
\end{equation}
where \(\theta = \{W_i, b_i\}_{i=1}^L\), \(W_i \in \mathbb{R}^{h_i \times h_{i-1}}\), \(b_i \in \mathbb{R}^{h_i}\), \(h_0 = n\), and \(h_L = 1\). Let the pre-activation variables be defined recursively as:
\begin{equation}
  z_0 = x, \quad
  z_i = W_i\,\sigma_i(z_{i-1}) + b_i, \quad
  z_L = h(x),
\end{equation}
where \(\sigma_i(\cdot)\) denotes the activation function at layer \(i\).

Using the chain rule, the Jacobian of the network output with respect to the input is given by:
\begin{equation}
  \nabla h(x) = \frac{\partial h}{\partial x}
  = W_1^\top J_1 W_2^\top J_2 \cdots W_{L-1}^\top J_{L-1} W_L^\top,
  \label{eq:jacobian}
\end{equation}
where each \(J_i = \mathrm{diag}\bigl(\sigma_i'(z_i)\bigr) \in \mathbb{R}^{h_i \times h_i}\) is a diagonal matrix containing the derivatives of the activation functions.

\subsection{Problem Formulation}
% \section{Problem Formulation}
% \label{sec:problem}
Having established these preliminaries, we now proceed to the Problem Formulation, where we formalize our objectives in certifying safety and outline the specific verification tasks.
The goal of this work is to formally verify the safety of the system \eqref{eq:dynamics} by certifying that the NCBFs condition \eqref{eq:cbf_condition} holds. The initial formulation of this problem checking the inequality for all points \( x \) within the continuous safe set \( C \) is computationally infeasible due to the infinite number of states. To simplify the problem, the first step is to reduce the verification domain. Based on Nagumo’s theorem~\cite{nagumo1942lage}, the condition only needs to be checked on the boundary of the safe set.

\begin{lemma}[Safety Verification on the Boundary]
\label{thm:verification_on_boundary}
Given a continuously differentiable NCBFs \( h(x) \), the safe set \( C := \{ x \mid h(x) \leq 0 \} \) for the control system \eqref{eq:dynamics} satisfy forward invariant condition~\eqref{eq:cbf_condition}, if the following inequality hold
\begin{equation}
    \mathcal{L}_f h(x) \leq -\alpha(h(x))=0, \quad \forall x \in \partial C, \quad \exists u \in \mathcal{U}.
    \label{eq:boundary_condition}
\end{equation}
where \( \alpha \geq 0 \) is a non-negative constant.
More generally, we incorporate \( \alpha \) to improve the **local attractiveness of the boundary** for discrete-time CBF-based safe control, especially when sequential system states cross the boundary.
Since \( h(x) = 0 \) on \( \partial C \), the class-\(\mathcal{K}\) term \( \alpha(h(x)) \) vanishes and the condition reduces to \( \mathcal{L}_f h(x) \leq 0 \), which is the necessary and sufficient condition for forward invariance from Nagumo’s theorem.
This lemma follows directly from \cite{nagumo1942lage} and is consistent with Theorem 1 in \cite{hu2024verification}.
\end{lemma}

\noindent\textbf{Remark:} In practice we take each subregion $\mathcal{R}_k$ to be an axis-aligned hyper-rectangle (convex, non-empty, and with non-zero Lebesgue measure). The covering $\bigcup_k\mathcal{R}_k$ may be conservative and subregions are allowed to overlap; non-overlap is not required for the correctness of Lemma~\ref{thm:verification_over_subregions}.

However, the boundary \( \partial C \) is the zero-level set of a complex neural network function. Precisely representing this high-dimensional, non-convex set is extremely difficult. To address this challenge, we replace the exact boundary with a conservative over-approximation composed of simple geometric shapes.

\begin{lemma}[Subregions Safety Verification]
\label{thm:verification_over_subregions}
Let the boundary \( \partial C \) be conservatively covered by a finite union of \( K \) subregions \( \{\mathcal{R}_k\}_{k=1}^K \), such that \( \partial C \subset \bigcup_{k=1}^K \mathcal{R}_k \). The system is certified safe if for each subregion \( \mathcal{R}_k \), the following holds:
\begin{equation}
    \mathcal{L}_f h(x) + \alpha(h(x)) \leq 0, \quad \forall x \in \mathcal{R}_k, \quad \exists\;u \in U.
    \label{eq:subregion_condition_robust}
\end{equation}
This condition is stronger than that of Lemma~\ref{thm:verification_on_boundary} and thus guarantees its satisfaction, since for any point on \( \partial C \cap \mathcal{R}_k \), the condition reduces to \eqref{eq:boundary_condition}.
\end{lemma}

By Lemma~\ref{thm:verification_over_subregions}, the safety verification problem reduces to checking whether, for each subregion \(\mathcal{R}_k\) covering the boundary \(\partial C\), there exists a control input \(u \in U\) such that condition~\eqref{eq:subregion_condition_robust} holds. If this is satisfied for all subregions, the system is certified safe.

\begin{figure*}[t]
  \centering
  \includegraphics[width=0.9\linewidth]{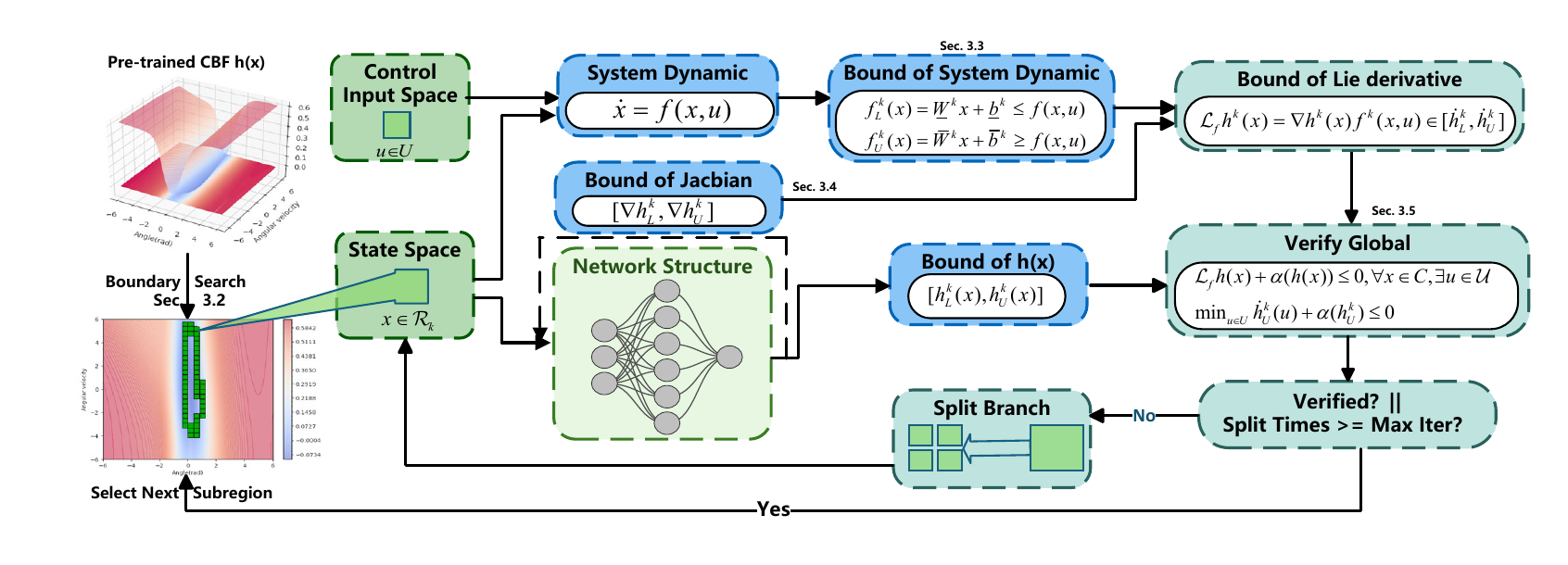}
  \caption{Overview of the verification pipeline: (1) search subregions (Section~\ref{sec:search_subregion}), (2) dynamics bounding (Section~\ref{sec:bound_sys}), (2) Jacobian bounding  (Section~\ref{sec:bound_ncbf}), (3) Lie‐derivative bound computation and verification condition check (Section~\ref{sec:verify_ncbf}).}
  \label{fig:overview}
\end{figure*}

\section{Verification Framework}
\label{sec:verification_framework}

Figure~\ref{fig:overview} illustrates our four-stage pipeline for enforcing forward invariance on each subregion $\mathcal{R}_k$,  similarly to~\cite{hu2024verification}. Starting from a pre-trained CBF, we first define the verification objective and necessary symbolic representations over subregions (Section~\ref{sec:Verif_obj_and_subregion}) and then conservatively cover the boundary $\partial C$ with a finite collection of subregions $\{\mathcal{R}_k\}$ (Section~\ref{sec:search_subregion}), following Lemma~\ref{thm:verification_over_subregions}. For each subregion, the CBF condition is then verified through a sequence of bounding and checking steps.

First, we bound the system dynamics \(f(x,u)\) by computing affine Taylor bounds \([f_L^k,f_U^k]\) (Section~\ref{sec:bound_sys}).
Next, we estimate the Jacobian bounds \([\nabla h_L^k,\nabla h_U^k]\) by propagating input intervals \([x_L^k,x_U^k]\) through the neural network and bounding the layer-wise derivative matrices \(\{[J_{i,L}^k,J_{i,U}^k]\}_{i=1}^{L-1}\) (Section~\ref{sec:bound_ncbf}).  
Then, we compute Lie‐derivative bounds \([\dot h_L^k,\dot h_U^k]\) via interval multiplication between the Jacobian and dynamics bounds (Section~\ref{sec:verify_ncbf}), and verify the condition \eqref{eq:subregion_minimax_relu} by checking whether \(\dot h_U^k+\alpha(h_U^k)\le0\) for \(u\in U\).  
Finally, if all subregions satisfy this condition, we certify that the CBF condition holds over the entire region.

\subsection{Verification objective and subregion notation}
\label{sec:Verif_obj_and_subregion}
By Lemma \ref{thm:verification_over_subregions}, safety verification on the boundary $\partial C$ reduces to checking a finite cover $\{\mathcal{R}_k\}_{k=1}^K$. 
To state the subregion-wise verification objective compactly, we introduce the following per-subregion notation.

Let
\[
\mathcal{R}_k = \{x\in D \mid x \in [x_L^k, x_U^k]\},
\]
where $x_L^k,x_U^k\in\mathbb{R}^n$ are the lower/upper state bounds on $\mathcal{R}_k$.  We also denote the interval bounds of the CBF itself by
\[
h(x)\in [h_L^k,h_U^k]\quad\text{for }x\in\mathcal{R}_k.
\]

Over $\mathcal{R}_k$ we characterize the system and network quantities by interval bounds:
\begin{itemize}
  \item Dynamics bounds: the dynamics $f^k(x,u)$ satisfy
  \[
  f^k(x,u)\in [f_L^k,f_U^k]\subset\mathbb{R}^n.
  \]
  \item Activation-derivative bounds: for layer $i$ the diagonal derivative matrix satisfies
  \[
  J_i^k=\mathrm{diag}(\sigma_i'(z_i))\in [J_{i,L}^k,J_{i,U}^k].
  \]
  where $\sigma_i'(\cdot)$ is the $i$-th layer activation function derivative, $z_i^k$ is the pre-activation input in $\mathcal{R}_k$
  \item Jacobian bounds of the neural CBF:
  \[
  \nabla h^k(x)=W_1^\top J_1^k W_2^\top\cdots W_{L-1}^\top J_{L-1}^k W_L^\top
  \in [\nabla h_L^k,\nabla h_U^k].
  \]
  \item Lie-derivative bounds:
  \[
  \mathcal{L}_f^k h(x) \coloneqq \nabla h^k(x)^\top f^k(x,u) \in [\dot h_L^k,\dot h_U^k].
  \]
\end{itemize}

% Within $\mathcal{R}_k$, we define interval-valued quantities to characterize local properties as follows. The system dynamics restricted to $\mathcal{R}_k$, denoted $f^k(x,u)$, satisfies the interval bound:  
% \[
% f^k(x,u) \in [f_L^k, f_U^k],
% \]  
% where $f_L^k, f_U^k \in \mathbb{R}^n$ are the dynamics bounds in $\mathcal{R}_k$.  

% For the $i$-th layer, the activation derivative matrix over $\mathcal{R}_k$ is defined as:  
% \[
% J_i^k = \mathrm{diag}\left(\sigma_i'(z_i^k)\right) \in [J_{i,L}^k, J_{i,U}^k],
% \]  
% where $\sigma_i'(\cdot)$ is the $i$-th layer activation function derivative, $z_i^k$ is the pre-activation input in $\mathcal{R}_k$, and $[J_{i,L}^k, J_{i,U}^k]$ is the interval bound of $J_i^k$.  

% The Jacobian of the Neural Network-based Control Barrier Function (NCBF) over $\mathcal{R}_k$, denoted $\nabla h^k(x)$, is expressed as:  
% \[
% \nabla h^k(x) = W_1^\top J_1^k W_2^\top J_2^k \cdots W_{L-1}^\top J_{L-1}^k W_L^\top \in [\nabla h_L^k, \nabla h_U^k],
% \]  
% where $[\nabla h_L^k, \nabla h_U^k]$ is its interval bound in $\mathcal{R}_k$.  

% The Lie derivative of $h$ along the system dynamics over $\mathcal{R}_k$ is defined as:  
% \[
% \mathcal{L}_f^k h(x) \coloneqq \nabla h^k(x)^\top f^k(x,u) \in [\dot{h}_L^k, \dot{h}_U^k],
% \]  
% where $[\dot{h}_L^k, \dot{h}_U^k] \subset \mathbb{R}$ is the interval bound of the Lie derivative in $\mathcal{R}_k$.  

Using these subregion intervals, the subregion-wise verification target becomes a scalar inequality that must hold for every $\mathcal{R}_k$. Concretely, we require the worst-case (maximal) Lie derivative over admissible controls to satisfy the CBF inequality:
\begin{equation}
\label{eq:subregion_minimax_restate}
\min_{u\in\mathcal{U}}\; \dot h_U^k + \alpha\bigl(h_U^k\bigr) \le 0.
\end{equation}
If \eqref{eq:subregion_minimax_restate} holds for all $k=1,\dots,K$, then by Lemma~\ref{thm:verification_over_subregions} the safe set $C$ is forward invariant.

For practical implementation, it is convenient to rewrite the left-hand side of \eqref{eq:subregion_minimax_restate} in terms of an explicit upper bound of the inner product $(\nabla h^k)^\top f^k$. 
The key observation is that the worst-case value of this inner product over the interval bounds 
$[\nabla h_L^k,\nabla h_U^k]$ and affine bounds $[f_L^k,f_U^k]$ is attained component-wise, since the inner product is separable across dimensions. 
To capture this, we decompose each vector into its positive and negative parts, 
$[v]_+:=\max\{v,0\}$ and $[v]_-:=\min\{v,0\}$. 
The maximum contribution of each component is then obtained by pairing the gradient and dynamics bounds according to their signs:  
 \begin{itemize}
     \item negative–negative pairs are maximized by taking the lower bounds ($[\nabla h_L^k]_-^\top [f_L^k]_-$);
     \item positive–negative pairs are maximized by combining the smallest positive gradient with the largest negative dynamics ($[\nabla h_L^k]_+^\top [f_U^k]_-$);
     \item and similarly for the remaining cases.
 \end{itemize}

By summing these component-wise maximizations, we obtain an explicit computable upper bound, leading to the equivalent formulation
\begin{multline}
    \min_{u\in\mathcal{U}}\; \{[\nabla h_U^k]_+^\top\,[f_U^k]_+
    + [\nabla h_L^k]_+^\top\,[f_U^k]_-
    + [\nabla h_U^k]_-^\top\,[f_L^k]_+ \\
    + [\nabla h_L^k]_-^\top\,[f_L^k]_-
    + \alpha(h_U^k)\} \le 0.
    \label{eq:subregion_minimax_relu}
\end{multline}
Equation~\eqref{eq:subregion_minimax_relu} is thus an equivalent, efficiently computable condition. 
If it holds for all subregions $\mathcal{R}_k$, then the CBF is certified safe under the specified control constraints.

\subsection{Search Subregions Containing the Boundary}
\label{sec:search_subregion}

To verify the safety of a neural CBF, we first identify a finite cover of the zero-level set \(\partial C = \{x \mid h(x) = 0\}\), following the same procedure as in~\cite{hu2024verification}. Assuming the CBF \(h(x)\) is continuous and approximately monotonic over small enough regions, we over-approximate the boundary using a grid of subregions. For each subregion, we check the signs of the CBF evaluated at its vertices. If all vertices yield strictly positive or strictly negative values, then the zero-crossing does not occur inside the region and it is discarded. Otherwise, if the vertex values have mixed signs, we retain the region as a candidate for verification. The detailed procedure is summarized in Algorithm~\ref{alg:search_boundary}. 

\begin{breakablealgorithm}
\caption{Search Subregions Containing the Boundary}
\label{alg:search_boundary}
\begin{algorithmic}[1]
\REQUIRE Gridded subregions \(\{\mathcal{X}_\text{sub}\}\), CBF function \(h(x)\)
\ENSURE Root-covering subregions \(\{\mathcal{R}_k\}_{k=1}^K\)

\STATE Initialize empty list \(\mathcal{R} = \emptyset\)
\FOR{each \(\mathcal{X}_\text{sub} \in \{\mathcal{X}_\text{sub}\}\)}
  \STATE Evaluate \(h(x)\) at all vertices \(v_i \in \mathcal{X}_\text{sub}\)
  \IF{\(\exists i,j\) such that \(h(v_i)\cdot h(v_j) < 0\)}
    \STATE Add \(\mathcal{X}_\text{sub}\) to \(\mathcal{R}\)
  \ENDIF
\ENDFOR
\RETURN \(\{\mathcal{R}_k\}_{k=1}^K := \mathcal{R}\)
\end{algorithmic}
\end{breakablealgorithm}

\subsection{Bounding System Dynamics Using Taylor Models}
\label{sec:bound_sys}
We assume that \(f(\cdot, u)\) is twice continuously differentiable in \(x\) for all admissible \(u\). Under these assumptions the Hessian-based Taylor remainder used below is well-defined and bounded (see \cite{hu2024real} for details).

Specifically, for each subregion \(\mathcal{R}_k\) with state bounds \(x \in [x_L^k, x_U^k]\), we linearize the dynamics about the mid point of the subregion $x^k_0$ and obtain affine bounds:
\begin{align*}
    f_L^k(x) = \underline{W}^k x + \underline{b}^k \leq f(x, u), \\
    f_U^k(x) = \overline{W}^k x + \overline{b}^k \geq f(x, u),
\end{align*}
where \(\underline{W}^k = \overline{W}^k = \nabla_x^\top f(x^k_0, u)\) and the residual terms \(\underline{b}^k, \overline{b}^k\) are computed entrywise as
\begin{align*}
    \underline{b}^k_i &= f_i(x^k_0, u) - (\nabla_x^\top f_i(x^k_0, u)) x^k_0 - \frac{1}{2} \| x_U^k - x_L^k \|_2^2 M_i, \\
    \overline{b}^k_i &= f_i(x^k_0, u) - (\nabla_x^\top f_i(x^k_0, u)) x^k_0 + \frac{1}{2} \| x_U^k - x_L^k \|_2^2 M_i,
\end{align*}
where \(M_i\) bounds the \(\ell_2\) operator norm of the Hessian of the \(i\)-th entry of \(f(x, u)\) within \(\mathcal{R}_k\), following \cite{hu2024real}. These affine bounds provide a generally valid lower and upper approximation of \(f(x, u)\) over the subregion.

\subsection{Bounding the Jacobian of the Neural CBF}
\label{sec:bound_ncbf}

To certify the properties of the NCBF, we must compute interval bounds for its Jacobian, denoted as $[\nabla h_L^k, \nabla h_U^k]$, over each region $\mathcal{R}_k$. The overall procedure for this is a backward propagation process that is common to both the original CROWN method and our proposed LightCROWN. The difference between the two approaches lies in a single, critical step: how they bound the derivatives of the activation functions $J_i$.

The Jacobian of the network, $h(x)$, can be expressed via the chain rule as $\nabla h(x) = J_1 W_2^\top J_2 \cdots J_{L-1}W_L^\top$, where $J_i = \text{diag}(\sigma_i'(z_i^k))$ is the diagonal matrix of activation derivatives at layer $i$. To compute interval bounds for this product, we employ the backward recursive procedure detailed in Algorithm~\ref{alg:jacobian_bounds}. This algorithm propagates interval bounds layer by layer, and finally returns $(Q_L^{(1)}, Q_U^{(1)})$, which correspond exactly to the lower and upper Jacobian bounds $[\nabla h_L^k, \nabla h_U^k]$. Here, the notation $[M]_+ = \max\{M,0\}$ and $[M]_- = \min\{M,0\}$ represents the element-wise positive and negative parts of a matrix $M$, respectively.
\begin{breakablealgorithm}
% \begin{algorithm}
\caption{General Backward Propagation for Jacobian Bounds}
\label{alg:jacobian_bounds}
\begin{algorithmic}[1]
\REQUIRE Network weights $\{W_j\}_{j=1}^L$; Interval bounds for activation derivatives $\{[J_{j,L}, J_{j,U}]\}_{j=1}^{L-1}$.
\ENSURE Interval bounds for the Jacobian $[\nabla h_L^k, \nabla h_U^k]$.
\STATE \textbf{Initialize:} $Q_L^{(L)} \leftarrow W_L^\top$ and $Q_U^{(L)} \leftarrow W_L^\top$.
\FOR{$j = L-1$ \textbf{down to} $1$}
    \STATE{Step 1: Multiply by the derivative bounds matrix $J_j$}
    \STATE $\widetilde Q_L^{(j)} \leftarrow J_{j,L}\,[Q_L^{(j+1)}]_+ + J_{j,U}\,[Q_L^{(j+1)}]_-$
    \STATE $\widetilde Q_U^{(j)} \leftarrow J_{j,U}\,[Q_U^{(j+1)}]_+ + J_{j,L}\,[Q_U^{(j+1)}]_-$
    \STATE{Step 2: Multiply by the weight matrix $W_j^\top$}
    \STATE $Q_L^{(j)} \leftarrow [W_j^\top]_+\,\widetilde Q_L^{(j)} + [W_j^\top]_-\,\widetilde Q_U^{(j)}$
    \STATE $Q_U^{(j)} \leftarrow [W_j^\top]_+\,\widetilde Q_U^{(j)} + [W_j^\top]_-\,\widetilde Q_L^{(j)}$
\ENDFOR
\RETURN $(Q_L^{(1)}, Q_U^{(1)})$
\end{algorithmic}
% \end{algorithm}
\end{breakablealgorithm}

As Algorithm~\ref{alg:jacobian_bounds} shows, the tightness of the final Jacobian bounds $[\nabla h_L^k, \nabla h_U^k]$ critically depends on the quality of the input derivative bounds $[J_{j,L}, J_{j,U}]$. We therefore turn to how these bounds are computed in CROWN and LightCROWN. The original CROWN framework takes an indirect route: it first bounds the activation outputs $\sigma_i(z_i^k)$ with linear upper and lower relaxations, and then infers derivative bounds from them. In contrast, LightCROWN directly leverages the analytical properties of  $\sigma_i'(z)$ over the interval $[z_{i,L}^k, z_{i,U}^k]$ to compute exact minimum and maximum values. For $(\sigma(z)=\tanh(z))$, the derivative is $\sigma_i'(z) = 1 - \sigma_i^2(z)$, the derivative attains its maximum at $z=0$ and its minimum at the endpoint with larger magnitude, leading to the following precise bounds:

\begin{equation} \label{eq:jacobian_bounds_lightcrown}
    (J_{i,L}^k,\,J_{i,U}^k) =\\
    \begin{cases}
        \bigl(\min\{\sigma_i'(z_{i,L}^k),\,\sigma_i'(z_{i,U}^k)\},\;1\bigr),\\
        \qquad \qquad \qquad \qquad \text{if } z_{i,L}^k \le 0 \le z_{i,U}^k, \\[6pt]
        \bigl(\sigma_i'(z_{i,L}^k),\;\sigma_i'(z_{i,U}^k)\bigr),
        \qquad \text{if } z_{i,U}^k \le 0, \\[4pt]
        \bigl(\sigma_i'(z_{i,U}^k),\;\sigma_i'(z_{i,L}^k)\bigr),
        \qquad \text{if } 0 \le z_{i,L}^k.
    \end{cases}
\end{equation}
By computing the true minimum and maximum of the derivative function, LightCROWN generates bounds $[J_{i,L}^k, J_{i,U}^k]$ that are provably tighter than or equal to those from original CROWN. This targeted improvement propagates through the backward recursion, yielding a strictly tighter final estimate for the Jacobian bounds $[\nabla h_L^k, \nabla h_U^k]$.

For general smooth nonlinear activations whose derivatives are not monotonic (e.g., sigmoid, Swish), the same principle applies by explicitly identifying all stationary points of the derivative function. Specifically, let $\phi'(z)$ denote the activation derivative, and suppose its critical points (solutions to $\phi''(z)=0$) are given by a finite set $\mathcal{Z}^\star$. Then, over any interval $[z_{i,L}^k, z_{i,U}^k]$, the extrema of $\phi'(z)$ must be attained at either the interval endpoints or the critical points within the interval. 

Therefore, the exact bounds can be computed as
\begin{equation}
(J_{i,L}^k, J_{i,U}^k)
=
\left(
\min_{z \in \mathcal{C}_i^k} \phi'(z),\;
\max_{z \in \mathcal{C}_i^k} \phi'(z)
\right),
\end{equation}
where
\begin{equation}
\mathcal{C}_i^k
=
\{z_{i,L}^k, z_{i,U}^k\}
\cup
\bigl(\mathcal{Z}^\star \cap [z_{i,L}^k, z_{i,U}^k]\bigr).
\end{equation}

This procedure effectively partitions the interval into regions where $\phi'(z)$ is monotonic, enabling tight evaluation of its extrema without relying on linear relaxation. In practice, for activations such as Swish, the set $\mathcal{Z}^\star$ contains only a small number of global critical points, making the computation efficient.

\subsection{Verification of System Safety}
\label{sec:verify_ncbf}

Building on the tight interval bounds for the system dynamics $\dot x = f(x,u)$ (Section~\ref{sec:bound_sys}) and the NCBFs Jacobian $\nabla h(x)$ (Section~\ref{sec:bound_ncbf}), we now establish system safety through global verification. The complete procedure, summarized in Algorithm~\ref{alg:ncbf_verify_detailed}, enforces the relaxed subregion condition from Lemma~\ref{thm:verification_over_subregions} on each hyper-rectangle $\mathcal R_k$, thereby certifying the forward invariance in \eqref{eq:cbf_condition}.  

For tractability, we assume $f(x,u)$ is \textbf{twice continuously differentiable in $x$}, with inputs constrained to a \textbf{convex polytope $\mathcal{U}$}. Such a convexity assumption, also adopted in prior verification frameworks~\cite{liu2023safe,hu2024verification}, ensures that the extrema of $f(x,u)$ over $\mathcal{U}$ lie at the vertices of $\mathcal{U}$, allowing the existential condition in Lemma~\ref{thm:verification_on_boundary} to be verified vertex-wise. Notably, this convexity assumption is introduced solely to simplify the invariance check and does not affect the Jacobian bounds computed by LightCROWN, ensuring its broader applicability.

\begin{breakablealgorithm}
\caption{Global Safety Verification}
\label{alg:ncbf_verify_detailed}
\begin{algorithmic}[1]
\REQUIRE State space \(\mathcal D\), control set \(\mathcal U\), network \(\theta\), class-$\mathcal{K}$ function  \(\alpha\), maximum number of splits \(S_{\max}\).
\ENSURE \texttt{SAFE} or \texttt{UNSAFE} together with the unverified subregion \(\mathcal{R}_{\text{fail}}\).

\STATE \textbf{Boundary Search:} Obtain subregions \(\{\mathcal R_k\}_{k=1}^K\) covering the CBF zero-level set \(\{x \mid h(x) = 0\}\) by applying the Algorithm~\ref{alg:search_boundary} in Section~\ref{sec:search_subregion} from discretized state space \(\mathcal D\).
\FOR{each region \(\mathcal R_k\) in the initial cover}
  \STATE Initialize queue \(Q \leftarrow \{\mathcal R_k\}\), split count \(s_k \leftarrow 0\)
  \WHILE{\(Q \neq \emptyset\)}
    \STATE Pop subregion \(\mathcal R_k^{(j)} \in Q\)
    \STATE \textbf{Forward propagation:} compute pre‐activation intervals 
    \(\{[z_{i,L}^k,z_{i,U}^k]\}_{i=1}^L\).
    \STATE \textbf{Activation derivative bounds:} for each layer \(i\), obtain 
    \([J_{i,L}^k,J_{i,U}^k]\) via Eq.~\eqref{eq:jacobian_bounds_lightcrown}.
    \STATE \textbf{Backward propagation:} compute Jacobian bounds 
    \([\nabla h_L^k,\nabla h_U^k]\) for 
    \(\nabla h(x)^k=W_1^\top J_1^k \cdots J_{L-1}^k W_L^\top\)
    via Algorithm~\ref{alg:jacobian_bounds}.
    \STATE Compute dynamics bounds \([f^k_L, f^k_U]\) using Section~\ref{sec:bound_sys}.
    \STATE \textbf{Existential check:} Iterate over all vertices of \(\mathcal{U}\) to find a control \(u\) satisfying condition~\eqref{eq:subregion_minimax_relu}.
    % \begin{equation*}
    % \begin{aligned}
    %   S = [\nabla h_U^k]_+^\top[f_U^k]_+ 
    %     &+ [\nabla h_L^k]_+^\top[f_U^k]_- + [\nabla h_U^k]_-^\top[f_L^k]_+\\
    %     &+ [\nabla h_L^k]_-^\top[f_L^k]_- 
    %     + \alpha\,h_U^k.
    % \end{aligned}
    % \end{equation*}
    \IF{such a \(u\) is found}
      \STATE Mark subregion \(\mathcal R^k\) as verified; continue
    \ELSIF{\(s_k < S_{\max}\)}
      \STATE Subdivide \(\mathcal R_k^{(j)}\) by bisecting each dimension at its midpoint, resulting in \(2^n\) smaller subregions \(\{\mathcal R_k^{(j\ell)}\}\)
      \STATE Enqueue all resulting subregions into \(Q\)
      \STATE Increment split count \(s_k \leftarrow s_k + 1\)
    \ELSE
      \STATE Set \(\mathcal{R}_{\text{fail}} \leftarrow \mathcal R_k^{(j)}\)
      \RETURN \texttt{UNSAFE}, \(\mathcal{R}_{\text{fail}}\)
    \ENDIF
  \ENDWHILE
\ENDFOR
\RETURN \texttt{SAFE}
\end{algorithmic}
\end{breakablealgorithm}

If the algorithm outputs \texttt{SAFE}, a valid control input $u$ exists for every subregion $\mathcal{R}_k$ such that the CBF condition is satisfied. By Lemma~\ref{thm:verification_over_subregions}, this guarantees forward invariance of the entire safe set $C$. If the algorithm returns \texttt{UNSAFE} along with the corresponding unverifiable subregion $\mathcal{R}_{\text{fail}}$, this region identifies where the NCBF safety condition is violated. Such counterexample regions can be directly leveraged to perform focused sampling and targeted optimization during NCBF training, guiding the neural network to rectify local safety violations and boost the verifiability and safety performance of the learned barrier function.

% \begin{remark}
% While LightCROWN is presented here for NCBFs verification, its applicability extends beyond this setting. By leveraging the analytical monotonicity of smooth activation derivatives to compute tight bounds (Equation~\eqref{eq:jacobian_bounds_lightcrown}), LightCROWN can also provide precise estimates of neural network properties such as Jacobian bounds and Lipschitz constants. These capabilities are valuable for tasks including sensitivity analysis, robustness verification, and stability assessment in broader learning-based systems.
% \end{remark}

\section{Experiments}
\label{sec:experiments}

We evaluate LightCROWN on two fronts: (i) comparative verification—benchmarking against baselines across heterogeneous robotic systems under varying safety margins (where larger $\alpha$ indicates a stricter safety margin); and (ii) scalability—ablation studies on boundary discretization and network size. Results for (i) appear in Section~\ref{subsec:alpha_performance}; details for (ii) are in Sections~\ref{subsec:boundary_partitioning} and~\ref{subsec:network_size_impact}.

\subsection{Experimental Settings}
\label{subsec:experimental_setup}

All experiments were conducted on a workstation with Intel Xeon Platinum 8375C CPU, and 512~GB RAM. We evaluate LightCROWN on three robotic systems by verifying their respective NCBFs: Inverted Pendulum(2D), Dubins Car(3D), and Planar Quadrotor(6D). The state space for each system is discretized into uniform grids with resolutions of 600, 50, and 6 grids/dim, respectively, and the resulting subregions serve as input for Algorithm~\ref{alg:search_boundary}.

The NCBFs are modeled using a two-layer $\tanh$-activated MLP, $h(x) = M_\theta(x)$. Benefiting from the strong approximation capability of neural networks, this shallow architecture is sufficient to represent the NCBFs of follow considered systems with state spaces up to $6$-dimensions. The network is defined as:$M_\theta(x) = W_2 \tanh(W_1 x + b_1) + b_2$,where $W_1: \mathbb{R}^{d_{\text{in}}} \to \mathbb{R}^{h_1}$ and $W_2: \mathbb{R}^{h_1} \to \mathbb{R}$. The input dimension $d_{\text{in}}$ corresponds to each system's dynamics $\{2,3,6\}$. The initial hidden units $h_1$ for the Pendulum, Car, and Quadrotor systems are 6, 64, and 8, respectively. To further assess scalability, we also vary $h_1$ across $\{8, 16, 64, 128\}$. Finally, the class-$\mathcal{K}$ function in verification objective Eq.~\eqref{eq:subregion_minimax_restate} and Eq.~\eqref{eq:subregion_minimax_relu} is instantiated as a linear function, $\alpha(h^k_U) = \alpha h^k_U$, for a positive constant $\alpha$.

\subsection{System Models}

We consider three representative nonlinear control systems, each system is paired with a pre-trained NCBFs, and safety verification is conducted over bounded state and input spaces.

\paragraph{Inverted Pendulum}
The Inverted Pendulum is modeled as a double integrator system moving in a 1D plane, with state $x = \begin{bmatrix} \theta \quad \dot{\theta} \end{bmatrix}^\top$ comprising position and velocity components, and control \(u\) directly influencing acceleration. Its continuous-time dynamics are given by a linear system:
\begin{equation}
\begin{aligned}
\dot{\theta} &= \ddot{\theta}\\
\ddot{\theta} &= (mGL\sin(\theta) +u - b\dot{\theta})/{J},
\end{aligned}
\end{equation} 
where $\theta$ is the tilt angle of the pendulum, $\dot{\theta}$ is the velocity of the tilt angle, $u$ represents the externally applied control input, $m$ is the mass of the pendulum, $L$ is the distance from the center of gravity of the rod to the axis of rotation, $J$ is the moment of inertia of the pendulum, and $b$ is the coefficient of friction.

\paragraph{Dubins Car}
The Dubins Car system models nonholonomic vehicle motion in a plane, with state \(x = (p_x, p_y, \phi)\) where \(\phi\) denotes heading, and control \(u = (\epsilon, \eta)\) representing linear and angular accelerations. The dynamics are:
\begin{equation}
\begin{aligned}
\dot{p}_x &= \cos(\phi) + \epsilon,\\
\dot{p}_y &= \sin(\phi) + \eta,\\
\dot{\phi} &= \omega,
\end{aligned}
\end{equation}
where \(\omega\) is implicitly controlled. 
%The state and input spaces are bounded by \(0 \leq p_x, p_y \leq 4\), \(0 \leq \phi \leq \pi\), and \(-1 \leq \epsilon, \eta \leq 1\), respectively. The unsafe set occupies the region \(1.5 \leq p_x \leq 2.5\), \(0 \leq p_y \leq 2\).

\paragraph{Planar Quadrotor}
The Planar Quadrotor models the dynamics of a quadcopter restricted to planar motion, with six-dimensional state \(x = (p_x, p_y, \theta, v_x, v_y, \omega)\) and control \(u = (F_1, F_2)\) representing thrusts from two propellers. Its equations of motion are:
\begin{equation}
\begin{aligned}
m\ddot{p}_x &= (F_1 + F_2) \sin\theta, \\
m\ddot{p}_y &= (F_1 + F_2) \cos\theta - mg, \\
J\ddot{\theta} &= \ell(F_2 - F_1)/2,
\end{aligned}
\end{equation}
where \(m\) and \(J\) denote the mass and moment of inertia, \(g\) is gravitational acceleration, and \(\ell\) is the distance between propellers. 
%The state constraints are \(0 \leq p_x, p_y \leq 4\), \(-0.1 \leq \theta \leq 0.1\), and \(-1 \leq v_x, v_y, \omega \leq 1\), while the control thrusts satisfy \(4 \leq F_1, F_2 \leq 6\). Unsafe states are defined similarly within specified positional and orientation bounds.

\subsection{Comparative Verification Across System Dynamic}
\label{subsec:alpha_performance}

Building on the setup in \ref{subsec:experimental_setup}, Tables~\ref{tab:regular_performance} and~\ref{tab:adversarial_performance} report verified rates (fraction of subregions $\mathcal{R}_k$ certified safe) and runtimes. We compare LightCROWN to two baselines: a CROWN-based implementation adapted from ~\cite{hu2024verification} and BBV~\cite{wang2024simultaneous}. To create a general CROWN linear-relaxation baseline, we removed ReLU-specific symbolic optimizations from the code of ~\cite{hu2024verification}. LightCROWN was then implemented on this baseline, exploiting the properties of continuously differentiable activations (e.g., $\tanh$) and their monotonicity to tighten output and Jacobian bounds. BBV is also included because it was used as a baseline in ~\cite{hu2024verification}. While BBV uses the same pre-activation bounding as CROWN, it differs in symbolic propagation and its treatment of the dynamics: BBV applies a Taylor-based global bound $(f_L^k,f_U^k)$ per subregion, whereas our LightCROWN and CROWN baseline use affine-in-$x$ relaxations. To assess robustness and align learning with verification, we train networks under both regular and adversarial training~\cite{madry2017towards}. Adversarial training exposes the CBF to worst-case state perturbations with Projeccted Gradient Descent~\cite{liu2023safe}, which improves forward-invariance and the fraction of certifiable subregions. Results ($\alpha\in\{0,0.1,0.2,0.5\}$) are reported for the Inverted Pendulum, Dubins Car, and Planar Quadrotor.

\begin{table}[ht]
\centering
\caption{Verification performance under regular training}
\label{tab:regular_performance}
\resizebox{0.9\linewidth}{!}{%
% \begin{threeparttable}
\begin{tabular}{cccccccccc}
\toprule
\multirow{2}{*}{$\alpha$} & 
\multirow{2}{*}{\textbf{Verifier}} & 
\multicolumn{2}{c}{\textbf{Inverted Pendulum}} & 
\multicolumn{2}{c}{\textbf{Dubins Car}} & 
\multicolumn{2}{c}{\textbf{Planar Quadrotor}} \\
\cmidrule(lr){3-4} \cmidrule(lr){5-6} \cmidrule(lr){7-8}
 & & \textbf{Rate (\%)} & \textbf{Time (s)} & \textbf{Rate (\%)} & \textbf{Time (s)} & \textbf{Rate (\%)} & \textbf{Time (s)} \\
\midrule
\multirow{3}{*}{0} 
 & CROWN & 91.1 & 62 & 83.4 & 112 & 48.8 & 425 \\
 & \textbf{LightCROWN} & \textbf{91.1} & \textbf{62} & \textbf{86.0} & \textbf{116} & \textbf{71.2} & \textbf{264} \\
 & BBV & 90.8 & 35 & 84.1 & 140 & 48.8 & 421 \\
\midrule
\multirow{3}{*}{0.1} 
 & CROWN & 91.1 & 53 & 50.5 & 168 & 13.2 & 481 \\
 & \textbf{LightCROWN} & \textbf{91.1} & \textbf{53} & \textbf{47.9} & \textbf{189} & \textbf{13.2} & \textbf{510} \\
 & BBV & 90.8 & 33 & 50.7 & 211 & 13.2 & 610 \\
\midrule
\multirow{3}{*}{0.2} 
 & CROWN & 91.1 & 53 & 44.9 & 182 & 13.2 & 497 \\
 & \textbf{LightCROWN} & \textbf{91.1} & \textbf{53} & \textbf{44.5} & \textbf{200} & \textbf{13.2} & \textbf{503} \\
 & BBV & 90.8 & 34 & 45.1 & 222 & 13.2 & 610 \\
\midrule
\multirow{3}{*}{0.5} 
 & CROWN & 91.0 & 33 & 42.5 & 188 & 13.0 & 508 \\
 & \textbf{LightCROWN} & \textbf{91.0} & \textbf{33} & \textbf{42.0} & \textbf{207} & \textbf{12.8} & \textbf{513} \\
 & BBV & 90.8 & 34 & 42.8 & 234 & 13.0 & 602 \\
\bottomrule
\end{tabular}%

% \begin{tablenotes}
% \footnotesize
% \item \textit{Note:} Cell format: CROWN/LightCROWN/BBV. Bold indicates best in category. 
% % \item $\downarrow$51.2\%: LightCROWN verification drop from $\alpha=0$ to $\alpha=0.5$
% % \item 1.92$\times$: LightCROWN rate advantage over CROWN at $\alpha=0$
% \end{tablenotes}
% \end{threeparttable}
}
\end{table}

% \subsubsection*{Regular Training Observations}
%For regularly trained models (Table~\ref{tab:regular_performance}), we have:
%\begin{itemize}
%    \item \textbf{Inverted Pendulum}: All methods maintain $\sim$91\% verified rate with BBV showing 43.5\% speed advantage at $\alpha$=0 (35s vs 62s for CROWN/LightCROWN)
%    \item \textbf{Dubins Car}: LightCROWN achieves peak verified rate (86.0\% at $\alpha$=0) but shows 51.2\% degradation at $\alpha$=0.5
%    \item \textbf{Planar Quadrotor}: LightCROWN demonstrates 1.48$\times$ higher verified rates than CROWN at $\alpha$=0 while maintaining 61\% faster computation
%\end{itemize}
For regularly trained models (Table~\ref{tab:regular_performance}), on the Inverted Pendulum, all methods achieve a verified rate of $\sim$91\%, where BBV shows a 43.5\% speed advantage at $\alpha=0$ (35s vs.~62s for CROWN/LightCROWN). For the Dubins Car, LightCROWN attains the peak verified rate (86.0\% at $\alpha=0$) but experiences a 44.0\% degradation at $\alpha=0.5$. On the Planar Quadrotor, LightCROWN surpasses CROWN with 1.48$\times$ higher verified rates at $\alpha=0$ while maintaining 61\% faster computation.

\begin{table}[ht]
\centering
\caption{Verification performance under adversarial training}
\label{tab:adversarial_performance}
\resizebox{0.9\linewidth}{!}{%
% \begin{threeparttable}
\begin{tabular}{cccccccccc}
\toprule
\multirow{2}{*}{$\alpha$} & 
\multirow{2}{*}{\textbf{Verifier}} & 
\multicolumn{2}{c}{\textbf{Inverted Pendulum}} & 
\multicolumn{2}{c}{\textbf{Dubins Car}} & 
\multicolumn{2}{c}{\textbf{Planar Quadrotor}} \\
\cmidrule(lr){3-4} \cmidrule(lr){5-6} \cmidrule(lr){7-8}
 & & \textbf{Rate (\%)} & \textbf{Time (s)} & \textbf{Rate (\%)} & \textbf{Time (s)} & \textbf{Rate (\%)} & \textbf{Time (s)} \\
\midrule
\multirow{3}{*}{0} 
 & CROWN & 99.2 & 28 & 75.6 & 96 & 37.5 & 1503 \\
 & \textbf{LightCROWN} & \textbf{99.3} & \textbf{27} & \textbf{90.5} & \textbf{80} & \textbf{92.8} & \textbf{903} \\
 & BBV & 99.2 & 48 & 75.7 & 84 & 42.1 & 1502 \\
\midrule
\multirow{3}{*}{0.1} 
 & CROWN & 99.2 & 28 & 75.5 & 98 & 29.4 & 1611 \\
 & \textbf{LightCROWN} & \textbf{99.3} & \textbf{38} & \textbf{88.5} & \textbf{84} & \textbf{87.3} & \textbf{1148} \\
 & BBV & 99.2 & 88 & 73.6 & 90 & 32.7 & 1298 \\
\midrule
\multirow{3}{*}{0.2} 
 & CROWN & 99.2 & 28 & 71.8 & 102 & 23.4 & 1771 \\
 & \textbf{LightCROWN} & \textbf{99.2} & \textbf{48} & \textbf{80.4} & \textbf{89} & \textbf{79.0} & \textbf{1196} \\
 & BBV & 99.2 & 79 & 73.6 & 91 & 25.6 & 1396 \\
\midrule
\multirow{3}{*}{0.5} 
 & CROWN & 99.0 & 28 & 67.2 & 107 & 13.6 & 1788 \\
 & \textbf{LightCROWN} & \textbf{99.2} & \textbf{38} & \textbf{80.4} & \textbf{99} & \textbf{44.7} & \textbf{1168} \\
 & BBV & 99.0 & 48 & 67.4 & 99 & 15.0 & 1570 \\
\bottomrule
\end{tabular}%
% \begin{tablenotes}
% \footnotesize
% \item \textit{Note:} Cell format: CROWN/LightCROWN/BBV. Bold indicates best in category.
% % \item $\uparrow$52.1\%: LightCROWN improvement at $\alpha=0.5$ vs regular training
% % \item 2.47$\times$: LightCROWN rate advantage at $\alpha=0$
% % \item 40\%: LightCROWN time reduction at $\alpha=0$
% \end{tablenotes}
% \end{threeparttable}
}
\end{table}

% \subsubsection*{Adversarial Training Enhancements}
%For adversarial training models (Table~\ref{tab:adversarial_performance}), we have:
%\begin{itemize}
%    \item \textbf{Verification Stability}: LightCROWN maintains $\geq$80\% verification for Dubins Car across all $\alpha$ values, reducing $\alpha$-sensitivity by 32.8\% compared to regular training
%    \item \textbf{Computation Efficiency}: For Planar Quadrotor at $\alpha$=0, LightCROWN achieves 2.47$\times$ higher verified rate than CROWN (92.8\% vs 37.5\%) with 40\% faster certification (903s vs 1503s)
%    \item \textbf{Method Superiority}: LightCROWN demonstrates best-in-class performance for 83\% of rate metrics and 58\% of time metrics across tested configurations
%\end{itemize}

With adversarial training (Table~\ref{tab:adversarial_performance}), LightCROWN's advantages become more pronounced. It demonstrates enhanced verification stability, maintaining a rate of $\geq$80\% for the Dubins Car across all $\alpha$ values, which reduces $\alpha$-sensitivity by 32.8\% compared to regular training. In terms of computational efficiency, for the Planar Quadrotor at $\alpha=0$, LightCROWN achieves a 2.47$\times$ higher verified rate than CROWN (92.8\% vs.~37.5\%) while being 40\% faster (903s vs.~1503s).

As \(\alpha\) increases, we observe a "reversal" in the verified rate under regular training, but not under adversarial training. This occurs because regular training causes LightCROWN to produce a tightly concentrated distribution of \(\dot h_U^k\) just below zero, while CROWN produces a broader distribution with some bounds extending further negative. Consequently, at \(\alpha=0\), LightCROWN achieves a higher verified rate than CROWN. However, when a positive margin \(\alpha>0\) is introduced, fewer LightCROWN bounds satisfy the verification condition~\eqref{eq:subregion_minimax_restate}, temporarily reducing its verified rate below that of CROWN. This reversal effect disappears under adversarial training, where worst-case perturbations shift all bounds further away from zero, preventing small increases in \(\alpha\) from inducing a reversal.

% \subsection{Effect of Number of Boundary Hyper-Rectangles}
\subsection{Impact of Boundary Discretization}
\label{subsec:boundary_partitioning}

\begin{figure}[htbp]
    \centering
    \includegraphics[width=0.9\linewidth]{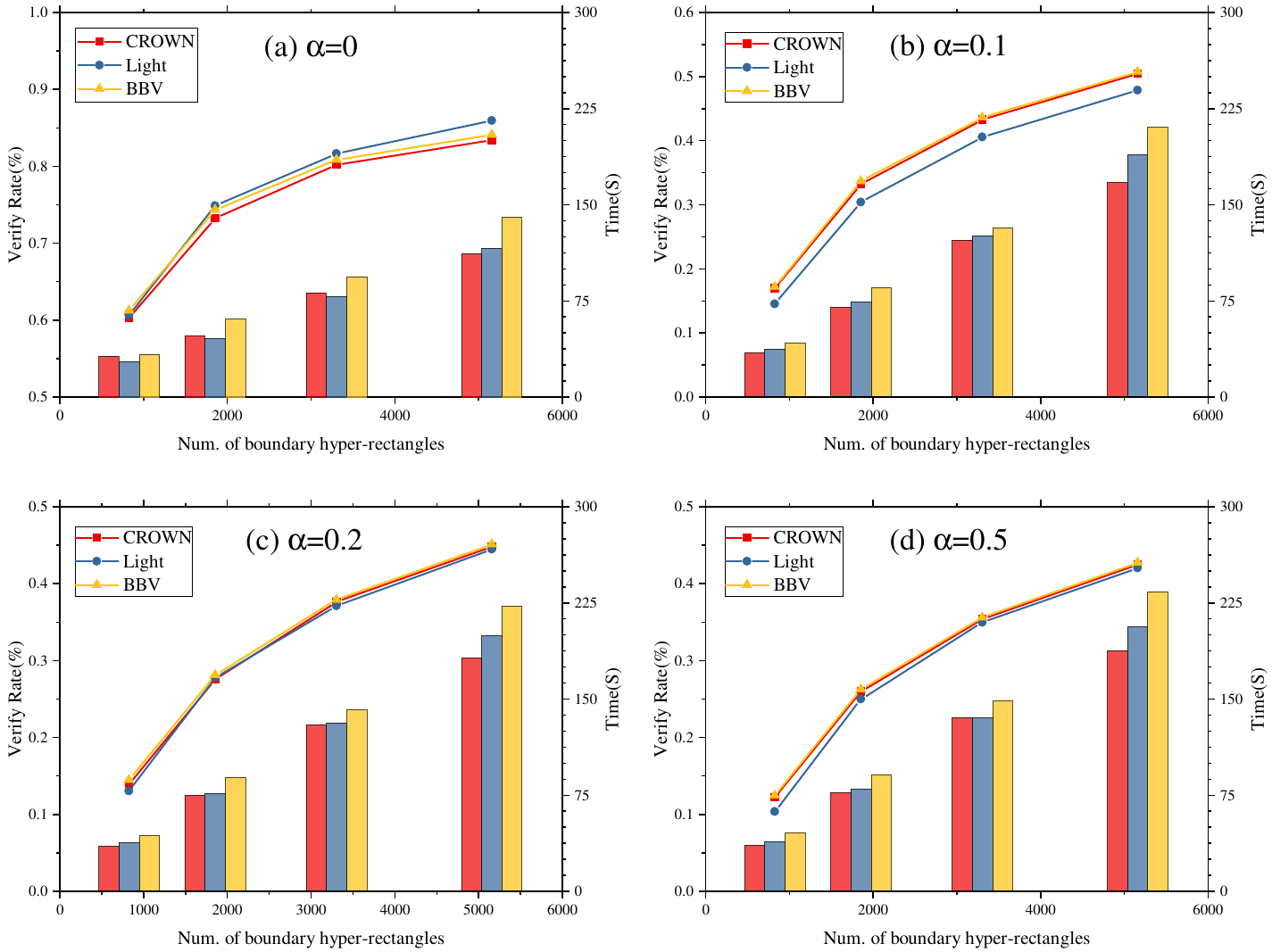}
    \caption{verified rate (lines, left axis) and verification time (bars, right axis) as a function of the number of subregions under regular training.  Each subplot corresponds to a different \(\alpha\in\{0,0.1,0.2,0.5\}\).}
    \label{fig:naive_and_hyper}
\end{figure}

\begin{figure}[htbp]
    \centering
    \includegraphics[width=0.9\linewidth]{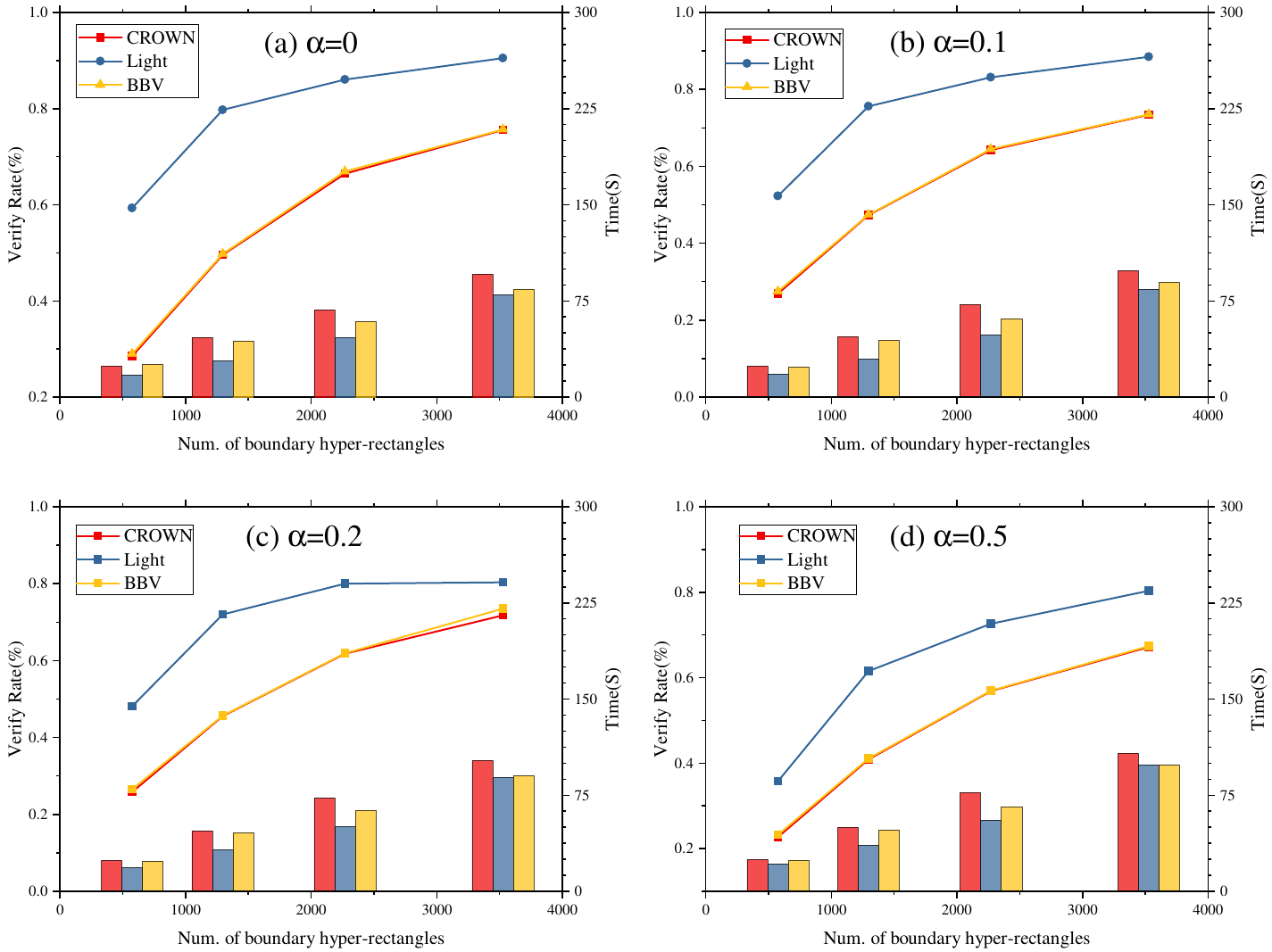}
    \caption{Same as the Fig.~\ref{fig:naive_and_hyper}, but under adversarial training.}
    \label{fig:adv_and_hyper}
\end{figure}

The above analysis demonstrates the efficiency of Light-CROWN, especially under adversarial training. To further assess its robustness, we evaluate its sensitivity to the granularity of the safety-boundary approximation. While finer subdivisions tighten per-subregion bounds, they also increase computational cost. To ensure that LightCROWN’s advantage is not tied to a specific setup, we vary the number of subregions and focus on the Dubins Car system with a 64-unit NCBF. Figures~\ref{fig:naive_and_hyper} and~\ref{fig:adv_and_hyper} report verified rates and computation times as the subregion count changes, using dual-axis plots (lines: verified rate; bars: computation time).

In the regular training case (Fig.~\ref{fig:naive_and_hyper}), LightCROWN initially outperforms both CROWN and BBV at low \(\alpha\), achieving higher verified rates and shorter runtimes.  However, as \(\alpha\) increases, we also observe a “reversal” in the verified rate: it falls below that of the baselines for \(\alpha=0.2\) and \(\alpha=0.5\), despite still running faster. 
% We hypothesize this reversal stems from the intrinsic “fragility” of regularly trained networks—many neurons operate on pre-activation intervals that span highly nonlinear regions of \(\tanh\), leading to wider propagated bounds in our direct-evaluation scheme.

By contrast, under adversarial training (Fig.~\ref{fig:adv_and_hyper}), the verified rates decrease monotonically with increasing $\alpha$, and LightCROWN consistently outperforms CROWN and BBV in both tightness and speed. As a result, LightCROWN achieves consistently higher verified rates across the entire range of $\alpha$, demonstrating its robustness for practical safety verification.

\subsection{Scalability with Network Capacity}
\label{subsec:network_size_impact}

To evaluate the scalability of our framework with respect to network size, we assessed the performance of CROWN, BBV, and LightCROWN on the Dubins Car model. The experiments varied the number of hidden-layer neurons across $\{8, 16, 64, 128\}$  while keeping the boundary discretization density fixed. The analysis was conducted for four different safety margins, \(\alpha\)\(\in\)\{0, 0.1, 0.2, 0.5\}.

\begin{figure}[ht]
    \centering
    \includegraphics[width=0.9\linewidth]{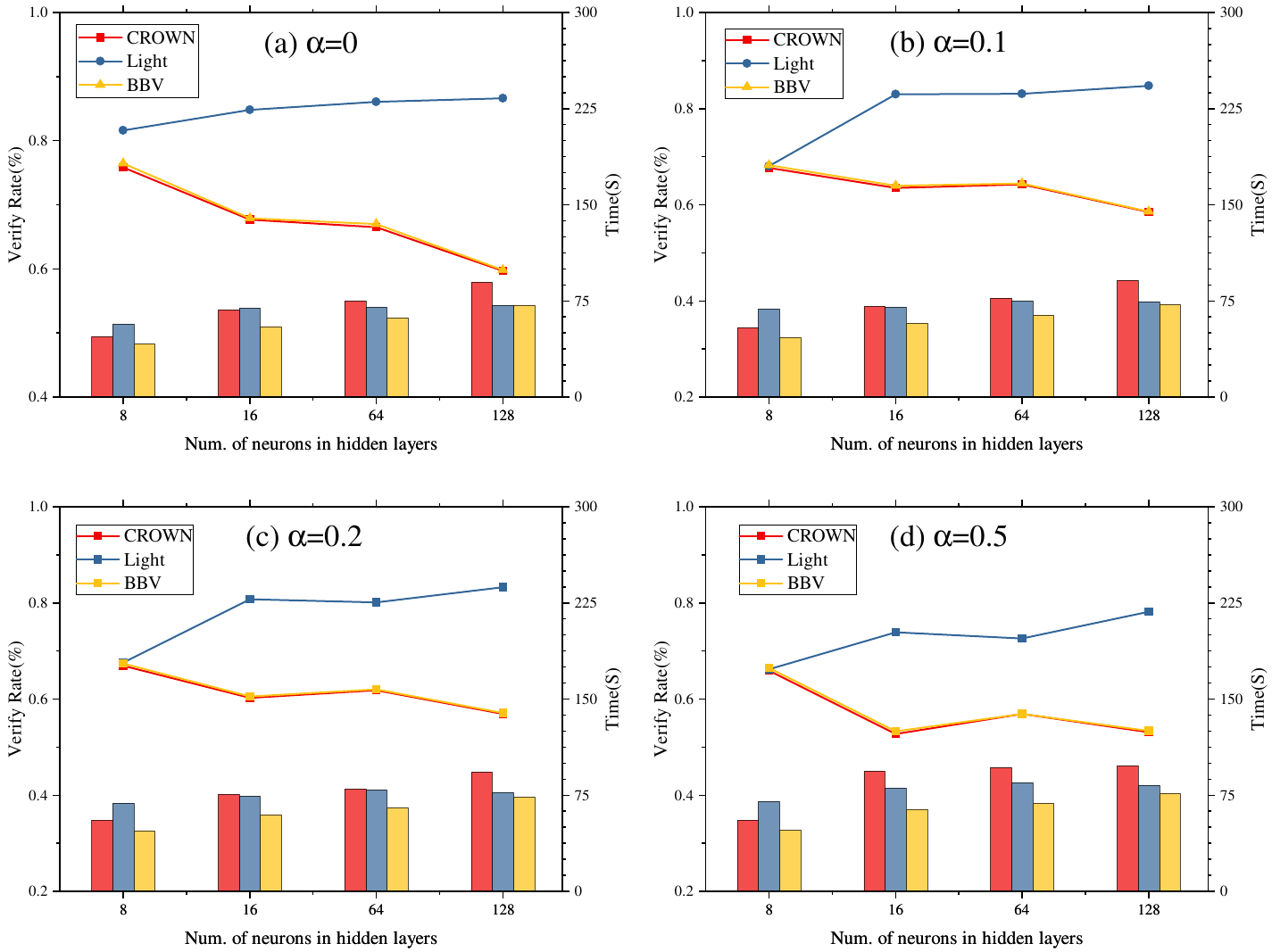}
    \caption{verified rate (lines, left axis) and verification time (bars, right axis)  for different hidden-layer widths for CROWN, BBV, and LightCROWN on the Dubins Car model. Each subplot corresponds to a value of \(\alpha\in\{0,0.1,0.2,0.5\}\).}
    \label{fig:impact_network_size}
\end{figure}

As illustrated in Figure~\ref{fig:impact_network_size}, while all methods show a modest decline in verified rate with increasing \(\alpha\), LightCROWN consistently outperforms the baselines across all network sizes. This advantage is particularly pronounced in larger networks. For instance, at \(\alpha\)=0.5 with 128 hidden neurons, LightCROWN certifies 78\% of the regions, significantly higher than CROWN (53\%) and BBV (53\%). Notably, BBV's performance degrades most sharply as both network size and \(\alpha\) increase, which underscores the robust scalability of our direct-bound strategy.

Overall, these results demonstrate that LightCROWN scales more effectively with network capacity, achieving substantially higher verified rates on larger and more complex neural networks. In terms of verification time, however, LightCROWN shows no clear advantage: depending on the setting, it can be slightly faster or slower than the baselines, but the differences remain relatively small across all tested configurations.

% \section{Discussion}

\section{Conclusion}
\label{sec:conclusion}

We presented LightCROWN, an efficient framework for formally verifying NCBFs with smooth, differentiable activations. The key contribution is a direct bounding technique that exploits the local monotonicity of activation derivatives to obtain significantly tighter Jacobian bounds than original CROWN-based methods. Experiments show that LightCROWN improves verified rates and scalability, and that adversarial training substantially enhances formal verifiability.

 % As observed in all tested systems except the simple inverted pendulum, achieving 100\% verification remains challenging. This limitation stems from inherent conservatism in interval-based verification, including: (i) over-approximation from Taylor models of nonlinear dynamics, (ii) relaxation from covering non-convex safety boundaries with subregions. LightCROWN reduces conservatism in derivative bounds but cannot eliminate all sources, yet it consistently achieves higher certification rates than prior methods under identical conditions.

Future directions include hybrid verification combining LightCROWN with precise solvers like SMT, where LightCROWN efficiently certifies most regions while the exact solver handles challenging cases. Incorporating higher-order dynamics models or advanced polynomial approximations could further tighten bounds, reduce conservatism, and enhance verifiability.

\bibliography{ref}

@article{nagumo1942lage,
  title={{\"U}ber die lage der integralkurven gew{\"o}hnlicher differentialgleichungen},
  author={Nagumo, Mitio},
  journal={Proceedings of the Physico-Mathematical Society of Japan. 3rd Series},
  volume={24},
  pages={551--559},
  year={1942},
  publisher={THE PHYSICAL SOCIETY OF JAPAN, The Mathematical Society of Japan}
}

@inproceedings{ames2019control,
  title={Control barrier functions: Theory and applications},
  author={Ames, Aaron D and Coogan, Samuel and Egerstedt, Magnus and Notomista, Gennaro and Sreenath, Koushil and Tabuada, Paulo},
  booktitle={2019 18th European control conference (ECC)},
  pages={3420--3431},
  year={2019},
  organization={IEEE}
}

@article{dawson2023safe,
  title={Safe control with learned certificates: A survey of neural lyapunov, barrier, and contraction methods for robotics and control},
  author={Dawson, Charles and Gao, Sicun and Fan, Chuchu},
  journal={IEEE Transactions on Robotics},
  volume={39},
  number={3},
  pages={1749--1767},
  year={2023},
  publisher={IEEE}
}

@article{hu2024verification,
  title={Verification of neural control barrier functions with symbolic derivative bounds propagation},
  author={Hu, Hanjiang and Yang, Yujie and Wei, Tianhao and Liu, Changliu},
  journal={arXiv preprint arXiv:2410.16281},
  year={2024}
}

@article{abate2020formal,
  title={Formal synthesis of Lyapunov neural networks},
  author={Abate, Alessandro and Ahmed, Daniele and Giacobbe, Mirco and Peruffo, Andrea},
  journal={IEEE Control Systems Letters},
  volume={5},
  number={3},
  pages={773--778},
  year={2020},
  publisher={IEEE}
}

@inproceedings{zhao2022verifying,
  title={Verifying neural network controlled systems using neural networks},
  author={Zhao, Qingye and Chen, Xin and Zhao, Zhuoyu and Zhang, Yifan and Tang, Enyi and Li, Xuandong},
  booktitle={Proceedings of the 25th ACM International Conference on Hybrid Systems: Computation and Control},
  pages={1--11},
  year={2022}
}

@article{tayal2024learning,
  title={Learning a Formally Verified Control Barrier Function in Stochastic Environment},
  author={Tayal, Manan and Zhang, Hongchao and Jagtap, Pushpak and Clark, Andrew and Kolathaya, Shishir},
  journal={arXiv preprint arXiv:2403.19332},
  year={2024}
}

@article{mathiesen2022safety,
  title={Safety certification for stochastic systems via neural barrier functions},
  author={Mathiesen, Frederik Baymler and Calvert, Simeon C and Laurenti, Luca},
  journal={IEEE Control Systems Letters},
  volume={7},
  pages={973--978},
  year={2022},
  publisher={IEEE}
}

@inproceedings{wang2024simultaneous,
  title={Simultaneous synthesis and verification of neural control barrier functions through branch-and-bound verification-in-the-loop training},
  author={Wang, Xinyu and Knoedler, Luzia and Mathiesen, Frederik Baymler and Alonso-Mora, Javier},
  booktitle={2024 European Control Conference (ECC)},
  pages={571--578},
  year={2024},
  organization={IEEE}
}

@article{wang2023simultaneous,
  title={Simultaneous synthesis and verification of neural control barrier functions through branch-and-bound verification-in-the-loop training},
  author={Wang, Xinyu and Knoedler, Luzia and Mathiesen, Frederik Baymler and Alonso-Mora, Javier},
  journal={arXiv preprint arXiv:2311.10438},
  year={2023}
}

@article{chen2024verification,
  title={Verification-Aided Learning of Neural Network Barrier Functions with Termination Guarantees},
  author={Chen, Shaoru and Molu, Lekan and Fazlyab, Mahyar},
  journal={arXiv preprint arXiv:2403.07308},
  year={2024}
}

@article{madry2017towards,
  title={Towards deep learning models resistant to adversarial attacks},
  author={Madry, Aleksander and Makelov, Aleksandar and Schmidt, Ludwig and Tsipras, Dimitris and Vladu, Adrian},
  journal={arXiv preprint arXiv:1706.06083},
  year={2017}
}

@inproceedings{liu2023safe,
  title={Safe control under input limits with neural control barrier functions},
  author={Liu, Simin and Liu, Changliu and Dolan, John},
  booktitle={Conference on Robot Learning},
  pages={1970--1980},
  year={2023},
  organization={PMLR}
}

@inproceedings{hu2024real,
  title={Real-Time Safe Control of Neural Network Dynamic Models with Sound Approximation},
  author={Hu, Hanjiang and Lan, Jianglin and Liu, Changliu},
  booktitle={6th Annual Learning for Dynamics \& Control Conference},
  year={2024},
  organization={PMLR}
}

@article{zhang2018efficient,
  title={Efficient neural network robustness certification with general activation functions},
  author={Zhang, Huan and Weng, Tsui-Wei and Chen, Pin-Yu and Hsieh, Cho-Jui and Daniel, Luca},
  journal={Advances in neural information processing systems},
  volume={31},
  year={2018}
}

@article{clark2024semi,
  title={A semi-algebraic framework for verification and synthesis of control barrier functions},
  author={Clark, Andrew},
  journal={IEEE Transactions on Automatic Control},
  year={2024},
  publisher={IEEE}
}

@article{vertovec2025scalable,
  title={Scalable Verification of Neural Control Barrier Functions Using Linear Bound Propagation},
  author={Vertovec, N. and Mathiesen, F. B. and Badings, T. and Laurenti, L. and Abate, A.},
  journal={arXiv preprint arXiv:2511.06341},
  year={2025}
}

@article{huang2025dynamic,
  title={Dynamic collision avoidance using velocity obstacle-based control barrier functions},
  author={Huang, Jihao and Zeng, Jun and Chi, Xuemin and Sreenath, Koushil and Liu, Zhitao and Su, Hongye},
  journal={IEEE Transactions on Control Systems Technology},
  year={2025},
  publisher={IEEE}
}

@inproceedings{gao2013dreal,
  title={dReal: An SMT solver for nonlinear theories over the reals},
  author={Gao, Sicun and Kong, Soonho and Clarke, Edmund M},
  booktitle={International conference on automated deduction},
  pages={208--214},
  year={2013},
  organization={Springer}
}

@inproceedings{peruffo2021automated,
  title={Automated and formal synthesis of neural barrier certificates for dynamical models},
  author={Peruffo, Andrea and Ahmed, Daniele and Abate, Alessandro},
  booktitle={International conference on tools and algorithms for the construction and analysis of systems},
  pages={370--388},
  year={2021},
  organization={Springer}
}

@article{xu2020automatic,
  title={Automatic perturbation analysis for scalable certified robustness and beyond},
  author={Xu, Kaidi and Shi, Zhouxing and Zhang, Huan and Wang, Yihan and Chang, Kai-Wei and Huang, Minlie and Kailkhura, Bhavya and Lin, Xue and Hsieh, Cho-Jui},
  journal={Advances in Neural Information Processing Systems},
  volume={33},
  pages={1129--1141},
  year={2020}
}

@article{yang2024lyapunov,
  title={Lyapunov-stable Neural Control for State and Output Feedback: A Novel Formulation for Efficient Synthesis and Verification},
  author={Yang, Lujie and Dai, Hongkai and Shi, Zhouxing and Hsieh, Cho-Jui and Tedrake, Russ and Zhang, Huan},
  journal={arXiv preprint arXiv:2404.07956},
  year={2024}
}

@article{ JQRR202303009,
author = { Chen, Mou and Ma, Haoxiang and Yong, Kenan and Wu Ying },
title = {Safety Flight Control of UAV: A Survey},
journal = {ROBOT},
volume = {45},
number = {03},
pages = {345-366},
year = {2023},
}

@article{ JQRR202501009,
author = { Fu, Junjie and Lin, Xiaokun and Wen, Guanghui },
title = {Robust Obstacle Avoidance and Safe Formation Tracking Control for Multiple Fixed-wing UAVs Based on High-order Control Barrier Functions},
journal = {ROBOT},
volume = {47},
number = {01},
pages = {85-98},
year = {2025},
}
\end{document}